\documentclass[runningheads]{llncs}
\usepackage{graphicx}

\usepackage{tikz}
\usepackage{comment} 
\usepackage{amsmath,amssymb} 
\usepackage{color}
\usepackage{multirow}

\begin{document}
\pagestyle{headings}
\mainmatter
\def\ECCVSubNumber{069}  

\title{AIM 2020 Challenge on Video Extreme Super-Resolution: Methods and Results} 

\titlerunning{AIM 2020 Challenge on Video Extreme Super-Resolution}
%
\author{Dario Fuoli\inst{1} \and
Zhiwu Huang\inst{1} \and
Shuhang Gu\inst{2} \and
Radu Timofte\inst{1}
Arnau Raventos\inst{3} \and Aryan Esfandiari\inst{3} \and Salah Karout\inst{3}  \and
Xuan Xu\inst{3}\and Xin Li\inst{3}\and Xin Xiong\inst{3}\and Jinge Wang\inst{3} \and
Pablo Navarrete Michelini\inst{3}\and Wenhao Zhang\inst{3} \and
Dongyang Zhang\inst{3}\and Hanwei Zhu\inst{3}\and Dan Xia\inst{3} \and
Haoyu Chen\inst{3} \and Jinjin Gu\inst{3} \and Zhi Zhang\inst{3} \and
Tongtong Zhao\inst{3} \and Shanshan Zhao\inst{3} \and
Kazutoshi Akita\inst{3} \and Norimichi Ukita\inst{3} \and
Hrishikesh P S\inst{3} \and Densen Puthussery\inst{3} \and Jiji C V\inst{3}}
\authorrunning{D. Fuoli et al.}
%
\institute{ETH Z\"urich, Switzerland \and University of Sydney, Australia \\
Dario Fuoli (\texttt{dario.fuoli@vision.ee.ethz.ch}), Zhiwu Huang, Shuhang Gu and Radu Timofte are the AIM 2020 challenge organizers. \url{http://www.vision.ee.ethz.ch/aim20/} \and
Participants in the challenge.
Appendix A contains the authors' teams and affiliations.
}
\maketitle

\begin{abstract}
This paper reviews the video extreme super-resolution challenge associated with the AIM 2020 workshop at ECCV 2020.
Common scaling factors for learned video super-resolution (VSR) do not go beyond factor 4. Missing information can be restored well in this region, especially in HR videos, where the high-frequency content mostly consists of texture details. The task in this challenge is to upscale videos with an extreme factor of 16, which results in more serious degradations that also affect the structural integrity of the videos. 
A single pixel in the low-resolution (LR) domain corresponds to 256 pixels in the high-resolution (HR) domain. Due to this massive information loss, it is hard to accurately restore the missing information. Track 1 is set up to gauge the state-of-the-art for such a demanding task, where fidelity to the ground truth is measured by PSNR and SSIM.
Perceptually higher quality can be achieved in trade-off for fidelity by generating plausible high-frequency content. Track 2 therefore aims at generating visually pleasing results, which are ranked according to human perception, evaluated by a user study.
In contrast to single image super-resolution (SISR), VSR can benefit from additional information in the temporal domain. However, this also imposes an additional requirement, as the generated frames need to be consistent along time.
\end{abstract}

\section{Introduction}

\begin{figure}[th!]
\begin{center}
\includegraphics[width=\linewidth]{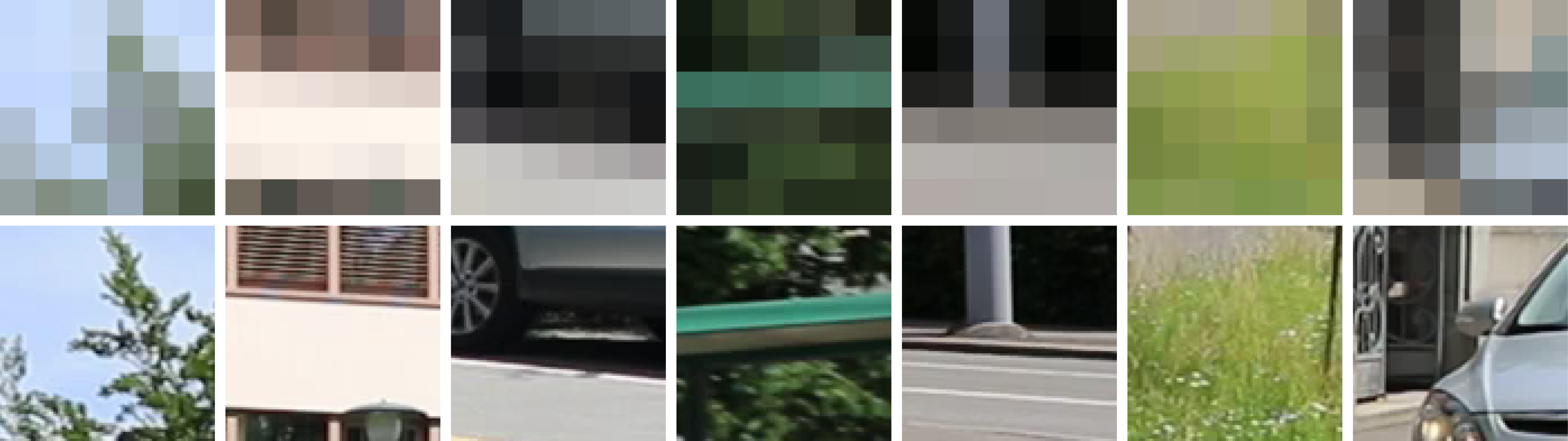}
\end{center}
   \caption{Downscaled crops with the extreme factor of $\times$16 (top) and corresponding 96$\times$96 HR crops (bottom).}
\label{fig:crops}
\end{figure}


Super-resolution (SR) aims at reconstructing a high-resolution (HR) output from a given low-resolution (LR) input. Single image SR (SISR) generally focuses on restoring spatial details as the input consists of only one single image. By comparison, as the input of video SR (VSR) is usually composed of consecutive frames, it is expected to concentrate on the exploitation of the additional temporal correlations, which can help improving restoration quality over SISR methods. Making full use of the temporal associations among multiple frames and keeping the temporal consistency for VSR remain non-trivial problems. Furthermore, when moving towards more extreme settings, like higher scale factors that require to restore a large amount of pixels from severely limited information, the VSR problem will get much more challenging. 

Following our first AIM challenge~\cite{AIM2019RVESRchallenge}, the goal of this challenge is to super resolve the given input videos with an extremely large zooming factor of 16, with searching for the current state-of-the-art and providing a standard benchmark protocol for future research in the field. Fig.\ref{fig:crops} presents a few downscaled crops and their corresponding HR crops. In this challenge, track 1 aims at probing the state-of-the-art for the extreme VSR task, where fidelity to the ground truth is measured by peak signal-to-noise ratio (PSNR) and structural similarity index (SSIM). As a trade-off for the fidelity measurement, track 2 is designed for the production of visually pleasing videos, which are ranked by human perception opinions with a user study. 


This challenge is one of the AIM 2020 associated challenges on:
scene relighting and illumination estimation~\cite{elhelou2020aim_relighting}, image extreme inpainting~\cite{ntavelis2020aim_inpainting}, learned image signal processing pipeline~\cite{ignatov2020aim_ISP}, rendering realistic bokeh~\cite{ignatov2020aim_bokeh}, real image super-resolution~\cite{wei2020aim_realSR}, efficient super-resolution~\cite{zhang2020aim_efficientSR}, video temporal super-resolution~\cite{son2020aim_VTSR} and video extreme super-resolution~\cite{fuoli2020aim_VXSR}.

\section{Related Work}

For SR, deep learning based methods ~\cite{srcnn,vdsr,srgan,enhancenet,photogan,dahl,lucas} have proven their superiority over traditional shallow learning methods. For example, \cite{srcnn} introduces convolution neural networks (CNN) to address the SISR problem. In particular, it proposes a very shallow network to deeply learn LR features, which are subsequently leveraged to generate HR images via non-linear mapping. To reduce the time complexity of the network operations in the HR space, \cite{ESPCN} proposes an effective sub-pixel convolution network to extract and map features from the LR space to the HR space using convolutional layers instead of classical interpolations (e.g., bilinear and bicubic). \cite{zhang2018residual} exploits a residual dense network block with direction connections for a more thorough extraction of local features from LR images.
A comprehensive overview of SISR methods can be found in \cite{timofte_challenge}.

Compared with SISR, the VSR problem is considerably more complex due to the additional challenge of harnessing the temporal correlations among adjacent frames. To address this problem, a number of methods~\cite{liao,endtoend,sparse_rep} are suggested to leverage temporal information by concatenating multiple LR frames to generate a single HR estimate. Following this strategy, ~\cite{caballero} first warps consecutive frames towards the center frame, and then fuses the frames using a spatio-temporal network. 
~\cite{kappeler} aggregates motion compensated adjacent frames, by computing optical flow and warping, followed by a few convolution layers for the processing on the fused frames. \cite{liu} calculates multiple HR estimates in parallel branches. In addition, it exploits an additional temporal modulation branch to balance the respective HR estimates for final aggregation. By
contrast, ~\cite{duf} relies on implicit motion estimation. Dynamic upsampling filters and residuals are computed from adjacent LR frames with a single neural network. Finally, the dynamic upsampling filters are employed to process the center frame, which is then fused with the residuals. Similarly, \cite{liu2020end} proposes a dynamic local filter network to perform implicit motion estimation and compensation. Besides, it suggests a global refinement neural network based on residual block and autoencoder structures to exploit non-local correlations and enhance the spatial consistency of the super-resolved frames. To address large motions, \cite{edvr} devises an alignment module to align frames with deformable convolutions in a coarse-to-fine manner. Besides, it suggests a fusion module, where attention is applied both temporally and spatially, so as to emphasize important features for subsequent restoration.

In addition to the aggregation strategy, some other works suggest to make use of recurrent neural networks (RNN) for better VSR. Due to the better capacity of learning temporal information on input frames, they provide a potentially more powerful alternative to address the SR problem. For instance, \cite{tao} suggests an autoencoder style network as well as an intermediate convolutional long short-term memory (LSTM) layer. The whole network is capable of processing the preliminary HR estimate from a subpixel motion compensation layer, for better HR estimate. \cite{brcn} proposes a bidirectional recurrent network, which exploits 2D and 3D convolutions with recurrent connections and combines a forward and a backward pass to produce the HR frames. To make use of temporal information, \cite{frvsr} designs a neural network that warps the previous HR output towards the current time step, by observing the optical flow in LR space. The warped output is concatenated with the current LR input frame and a SR network generates the HR estimate. \cite{rlsp} exploits a recurrent latent space propagation (RLSP) algorithm for more efficient VSR. Particularly, RLSP introduces high-dimensional latent states to propagate temporal information along frames in an implicit manner so that the efficiency can be highly improved.

\section{AIM 2020 Video Extreme SR Challenge Setup}

\subsection{Data}

\begin{figure*}[t]
\begin{center}
\includegraphics[width=\textwidth]{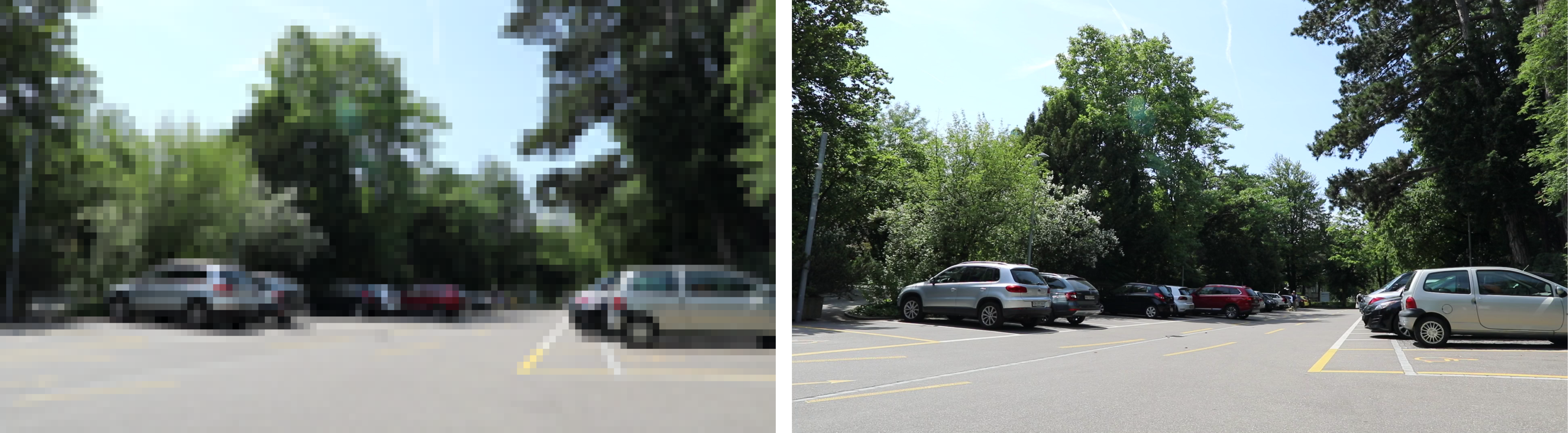}
\end{center}
   \caption{LR frame with extreme downscaling factor $\times$16 (left) and corresponding HR frame (right).}
\label{fig:global_lr_hr}
\end{figure*}

We use the Vid3oC \cite{aim_dataset} dataset for this challenge, which has been part of previous challenges \cite{AIM2019RVESRchallenge,NTIRE2020VQM}. The dataset is a collection of videos taken with three different cameras on a rig. This results in roughly aligned videos, which can be used for weak supervision. In this challenge, only the high-quality DSLR camera (Canon 5D Mark IV) is used to serve as HR ground truth. 
The corresponding LR source data is obtained by downscaling the ground truth by factor 16, using MATLAB's imresize function with standard settings, see Fig.~\ref{fig:global_lr_hr}. In order to retain proper pixel-alignment, the ground truth $1080\times1920$ FullHD frames are cropped to $1072\times1920$ before downscaling, to be dividable by 16.
We provide 50 HR sequences to be used for training. To save bandwidth, these videos are provided as MP4 files together with scripts to extract and generate the LR source frames. Additionally, the dataset contains 16 paired sequences for validation and 16 paired sequences for testing, each consists of 120 frames in PNG format.

\subsection{Challenge Phases}
The challenge is hosted on CodaLab and is split up in a validation and a test phase. During the validation phase, only the validation source frames are provided and participants were asked to submit their super-resolved frames to the CodaLab servers to get feedback. Due to storage constraints on CodaLab, only a subset of frames could be submitted to the servers (every 20th frame in the sequence).
In the following test phase, the final solutions had to be submitted to enter the challenge ranking. There was no feedback provided at this stage, in order to prevent overfitting to the test set. Additionally, the full set of frames had to be made accessible to the challenge organizers for the final rankings.
After the submission deadline, the HR validation ground truth was released on CodaLab, for public use of our dataset.


\subsection{Track 1 - Fidelity}
This track aims at high fidelity restoration. For each team, the restored frames are compared to the ground truth in terms of PSNR and SSIM and can be objectively quantified by these pixel-level metrics. The focus is on restoring the data faithfully to the underlying ground truth. Commonly, methods for this task are trained with a pixel-level L1-loss or L2-loss. The final ranking among teams is determined by PSNR/SSIM exclusively, without visual assessment of the produced frames.

\subsection{Track 2 - Perceptual}
Super-resolution methods optimized for PSNR tend to oversmooth and often fail to restore the highest frequencies. Also, PSNR does not correlate well with human perception of quality. Therefore, the focus in the field has shifted towards generation of perceptually more pleasing results in trade-off for fidelity to the ground truth. Since the extreme scaling factor 16 and its associated large information loss prohibits high fidelity results, the only possibility to achieve realistically looking HR videos in this setting, is by hallucinating plausible high frequencies. Track 2 is aimed at upscaling the videos for highest perceptual quality. Quantitative assessment of perceptual quality is difficult and remains largely an open problem. We therefore resort to a user study in track 2, which is still the most reliable benchmark for perceptual quality evaluation.

\section{Challenge Methods and Teams}


\subsection{KirinUK}

\begin{figure}[t]
\begin{center}
\includegraphics[width=1\linewidth]{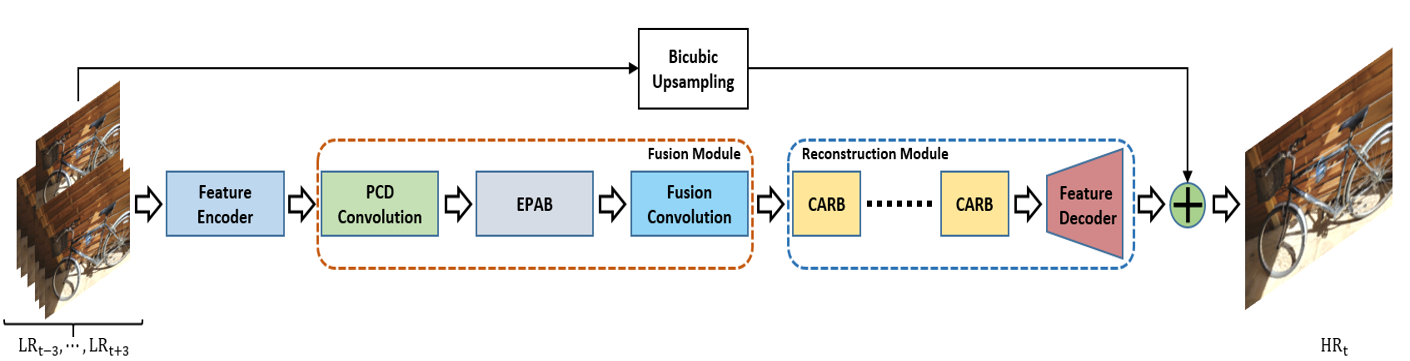}
\end{center}
   \caption{Efficient  Video  Enhancement  and  Super-Resolution  Net  (EVESRNet) proposed by KirinUK.}
\label{fig:kirinukarch}
\end{figure}

\begin{figure}[t]
\begin{center}
\includegraphics[width=0.65\linewidth]{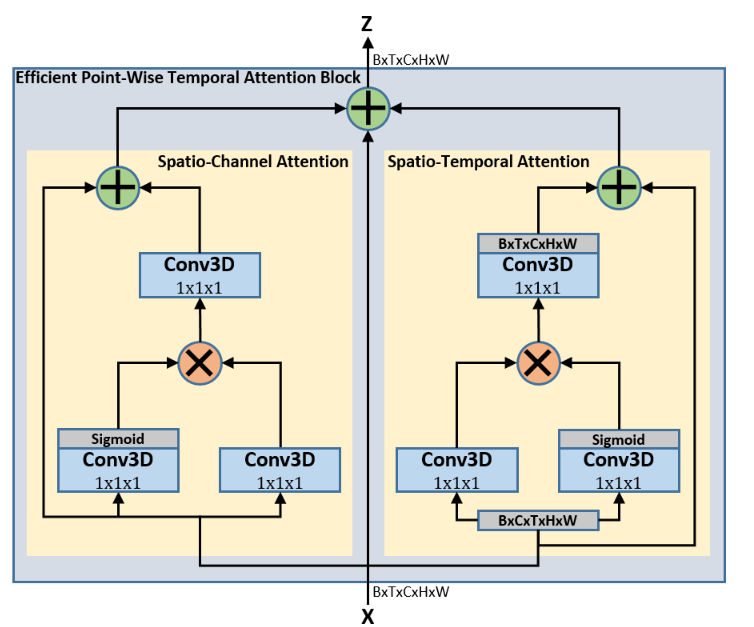}
\end{center}
   \caption{Efficient Point-Wise Temporal Attention Block proposed by KirinUK.}
\label{fig:kirinuk}

\end{figure}

Recent video super-resolution approaches~\cite{chen2020vesr} propose splitting the spatio-temporal attention operation in several dimensions. Their aim is to reduce the computational cost of a traditional 3D non-local block. Nevertheless, these methods still need to store the HWxHW attention matrices, which is challenging, especially when dealing with GPUs with limited amount of memory or when upscaling HR videos. To tackle this, the KirinUK team proposes to extend the VESRNet~\cite{chen2020vesr} architecture by replacing the Separate Non Local (SNL) module with an Efficient Point-Wise Temporal Attention Block (EPAB). This block aggregates the spatio-temporal information  with less operations and memory consumption, while still keeping the same performance. The team names this new architecture Efficient Video Enhancement and Super-Resolution Net (EVESRNet) and an overview of it can be seen in Fig.~\ref{fig:kirinukarch}. It is mainly composed of Pyramid, Cascading and Deformable Convolutions (PCDs)~\cite{edvr}, the EPAB, and Channel Attention Residual Blocks (CARBs)~\cite{chen2020vesr}~\cite{zhang2018image}.

The EPAB module is illustrated in Fig.\ref{fig:kirinuk} and can be divided into two sub-blocks, the Spatio-Channel Attention (SCA) and the Spatio-Temporal Attention (STA). Both of them share the same structure, the only difference is the permutation operation at the beginning and at the end of the STA sub-block. 

To perform Extreme Video SR, the team employs two 4x stages in cascade mode. Each stage was trained independently with an EVESRNet architecture. Moreover, their training does not start from scratch, they first pretrain a model with the REDS dataset~\cite{Nah_2019_CVPR_Workshops_REDS} to initialize the networks. This helps preventing overfitting in the first stage where the amount of spatial data is limited. The REDS dataset was only utilized for pretraining. As a reference, the initialization model achieves 31.19 PSNR in the internal REDS validation set defined in~\cite{edvr}, which corresponds to a +0.1 dB improvement with respect to EDVR~\cite{edvr}.

Each track uses the same pipeline, the only differences are: 1) For the fidelity track, the two stages were trained using L2 loss. 2) For the perceptual track, the second stage was trained using the following combined loss: 
\begin{equation}
L=\lambda_1*L_{L1} + \lambda_2*L_{VGG} + \lambda_3*L_{RaGAN}
\end{equation}
where $\lambda_1$ is 1e-3, $\lambda_2$ is 1 and $\lambda_3$ is 5e-3. They used a patch discriminator as in \cite{ji2020real}. The rest of the hyperparameters are the same as in \cite{wang2018esrgan}.

\subsection{Team-WVU}

\begin{figure}[t]
  \centering
  \includegraphics[width=0.85\linewidth]{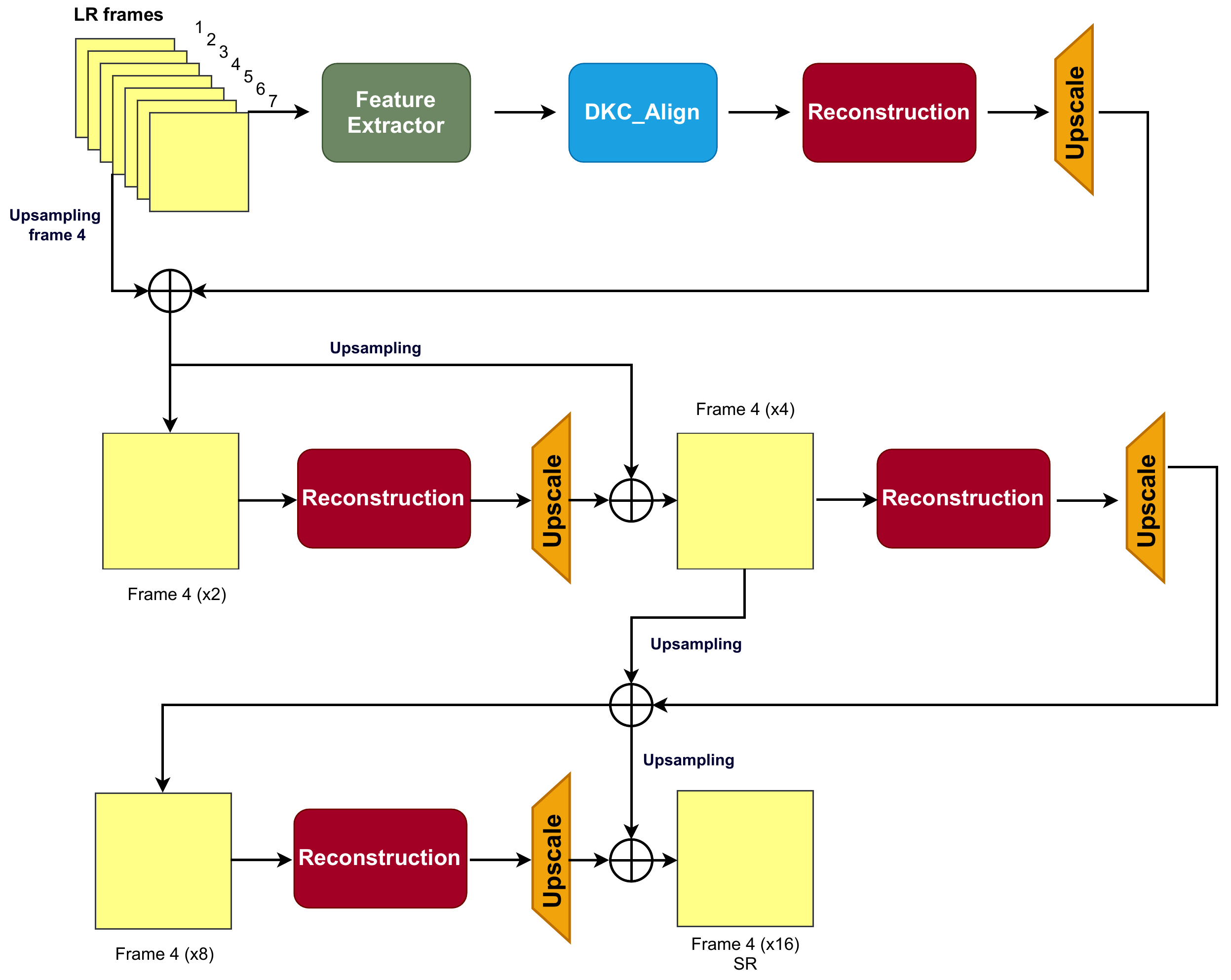}
  \caption{Network proposed by Team-WVU.}
  \label{fig:wvu_network}
\end{figure}

Recently, deformable convolution \cite{zhu2019deformable} has been received increasingly more attention to solve low-level vision tasks such as video super-resolution. EDVR \cite{edvr} and TDAN \cite{tian2020tdan} have already successfully implemented deformable convolution to temporally align reference frame and its neighboring frames which can let networks better utilize both spatial and temporal information to enhance the final results. 

Inspired by state-of-the-art video SR method EDVR \cite{edvr}, the team develops the novel Multi-Frame based Deformable Kernel Convolution Network \cite{xuan2020aim_vxsr} to temporally align the non-reference and reference frames with deformable kernel \cite{Gao2020Deformable} convolution alignment module and enhance the edge and texture features via deformable kernel spatial attention module. 

The overall diagram of proposed network is shown in Fig.~\ref{fig:wvu_network}. It mainly includes four parts, feature extractor, DKC\_Align module (deformable kernel convolution alignment module), reconstruction module and upscale module. Different from PCD alignment module from EDVR, the team implemented stacked deformable kernel convolution layers instead of traditional convolution layer to extract offset. Deformable kernel can better adapt effective receptive fields than normal convolution \cite{Gao2020Deformable} which can better enhance the offset extraction compared with the normal convolution. On the step of reconstruction, to calibrate reconstructed feature maps before feeding into each upscaling module, the team proposes a Deformable Kernel Spatial Attention (DKSA) module (integrated to reconstruction module) to enhance the textures that can help the proposed network to reconstruct SR frames sharper and clearer. Because this challenge aims to super-resolve extreme LR videos with the scale factor of 16, to avoid generating undesired blurring and artifacts, the LR frames are super-resolved with a scale factor of 2 each time (see Fig.~\ref{fig:wvu_network}). Finally, the LR frames are super-resolved four times in total to upscale the LR frames with a magnification factor $\times$16. Charbonnier Loss \cite{edvr,lai2017deep} is used as the loss function for both track 1 and track 2 training, the loss can be expressed as follows:
\begin{equation}
    L = \sqrt{||\hat{X}_r - X_r||^2+\xi^2}
\label{eqn:char_loss}
\end{equation}
where $\xi = 1\times 10^{-3}$, $\hat{X}_r$ is super-resolved frame and $X_r$ is target frame (ground-truth).

\subsection{BOE-IOT-AIBD}

\begin{figure}[t]
  \centering
  \includegraphics[width=0.8\linewidth]{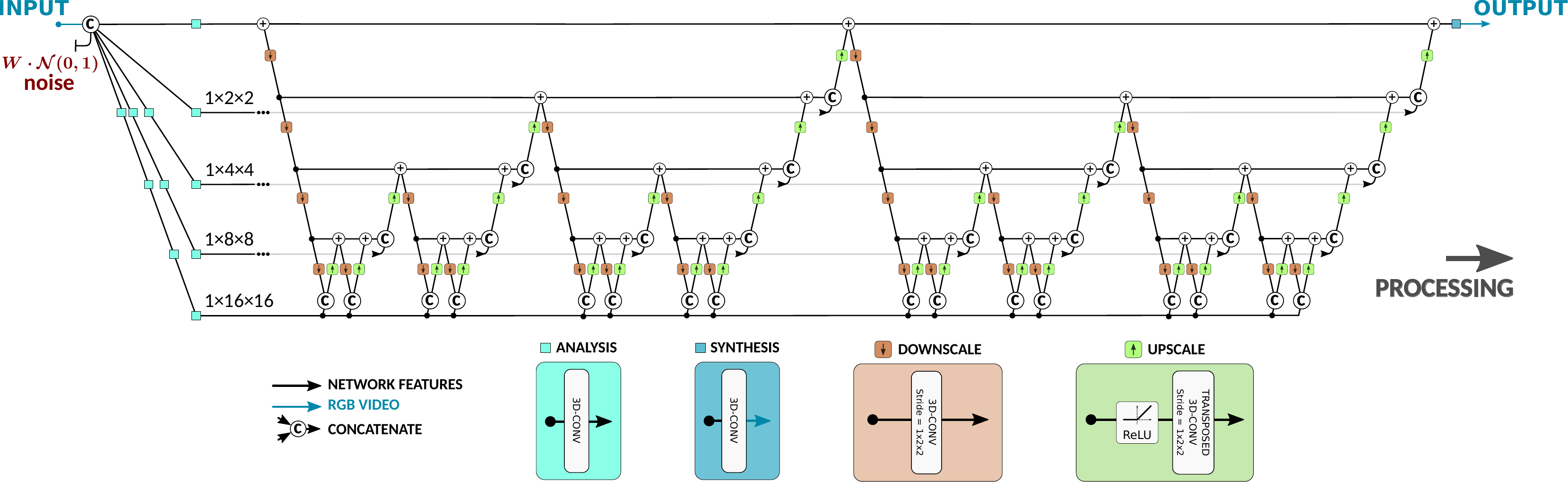}\\
  (a) 3D--MGBP Network
  \includegraphics[width=0.8\linewidth]{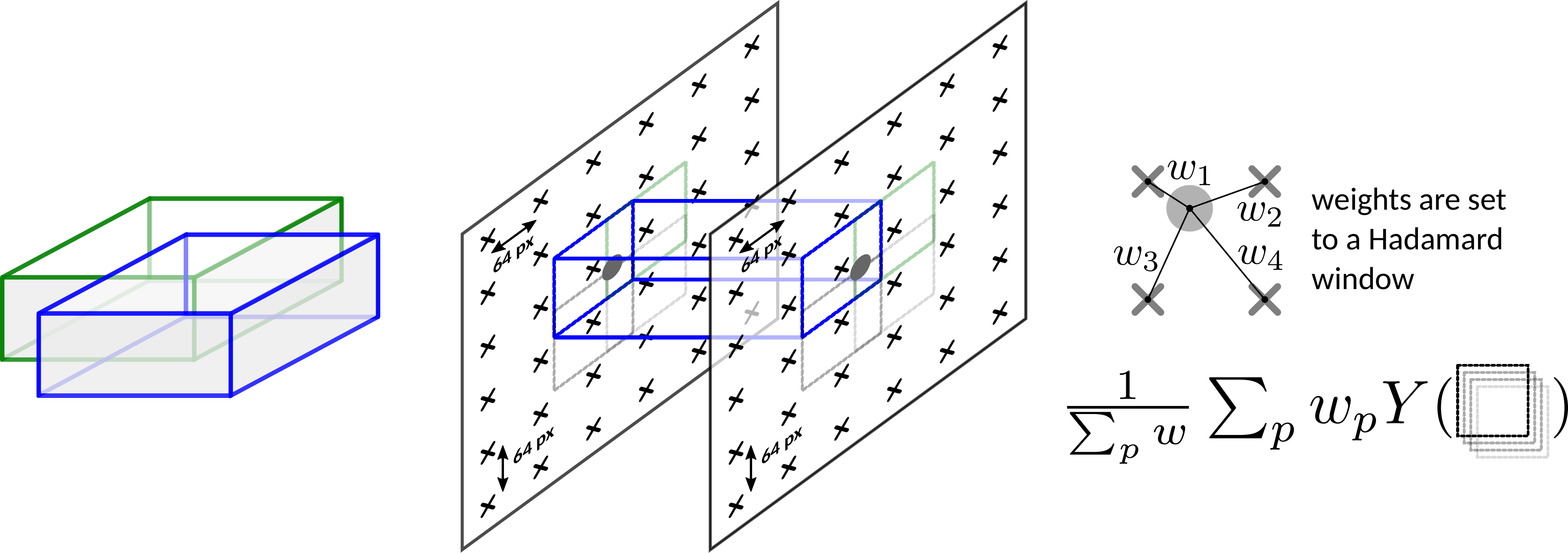}\\
  (b) Overlapping strategy

  \caption{3D -- MultiGrid BackProjection network (3D--MGBP) proposed by BOE-IOT-AIBD.}
  \label{fig:mgbp3d_boe}
\end{figure}

The team proposes 3D-MGBP, a fully 3D--convolutional architecture designed to scale efficiently for the difficult task of extreme video SR. 3D--MGBP is based on the Multi--Grid Back--Projection network introduced and studied in \cite{PNavarrete_2019a,MGBPv2,G-MGBP}. In particular, they extend the MGBPv2 network~\cite{MGBPv2} that was designed to scale efficiently for the task of extreme image SR and was successfully used in the 2019--AIM Extreme Image SR competition~\cite{lugmayr2019aim} to win the Perceptual track of that challenge. For this challenge they redesigned the MGBPv2 network to use 3D--convolutions strided in space. The network works as a video enhancer that, ignoring memory constrains, can take a whole video stream and outputs a whole video stream with the same resolution and framerate. They input a $16\times$ Bicubic upscaled video and the network enhances the quality of the video stream. The receptive field of the network extends in space as well as time by using 3D--convolutional kernels of size $3\times3\times3$. The overall architecture uses only 3D--convolutions and ReLU units. This is in contrast to general trends in video processing networks that often include attention, deformable convolutions, warping or other non--linear modules.

Figure \ref{fig:mgbp3d_boe} displays the diagram of the 3D--MGBP network used in the competition. In inference it is impossible for 3D--MGBP to process the whole video stream and so they extend the idea of overlapped patches used in MGBPv2 by using overlapped spatio--temporal patches (overlapping in space and time). More precisely, to upscale arbitrarily long video sequences they propose a patch based approach in which they average the output of overlapping video patches produced by the Bicubic upscaled input. First, they divide input streams into overlapping patches (of same size as training patches) as shown in Figure \ref{fig:mgbp3d_boe}; second, they multiply each output by weights set to a Hadamard window; and third, they average the results.

They trained the 3D--MGBP network starting from random parameters (no pre--trained models were used). For the Fidelity track they trained the model using L2 loss on the output spatio--temporal patch. For the Perceptual track they submitted the output of two different configurations of the same architecture. The first submission, labeled \emph{Smooth}, was trained with L2 loss as they noticed better time--consistency and smooth edges. In their second submission, labeled \emph{Texture}, they followed the loss and training strategy of G--MGBP~\cite{G-MGBP}, adding a noise input to activate and deactivate the generation of artificial details. The noise input consists of one channel of Gaussian noise concatenated to the Bicubic upscaled input. In this solution, although more noisy and farther from ground truth due to the perception--distortion trade--off \cite{Blau_2018_CVPR}, they noticed better perception of textures.

\subsection{ZZX}

\begin{figure}[t]
  \centering
  \includegraphics[width=\linewidth]{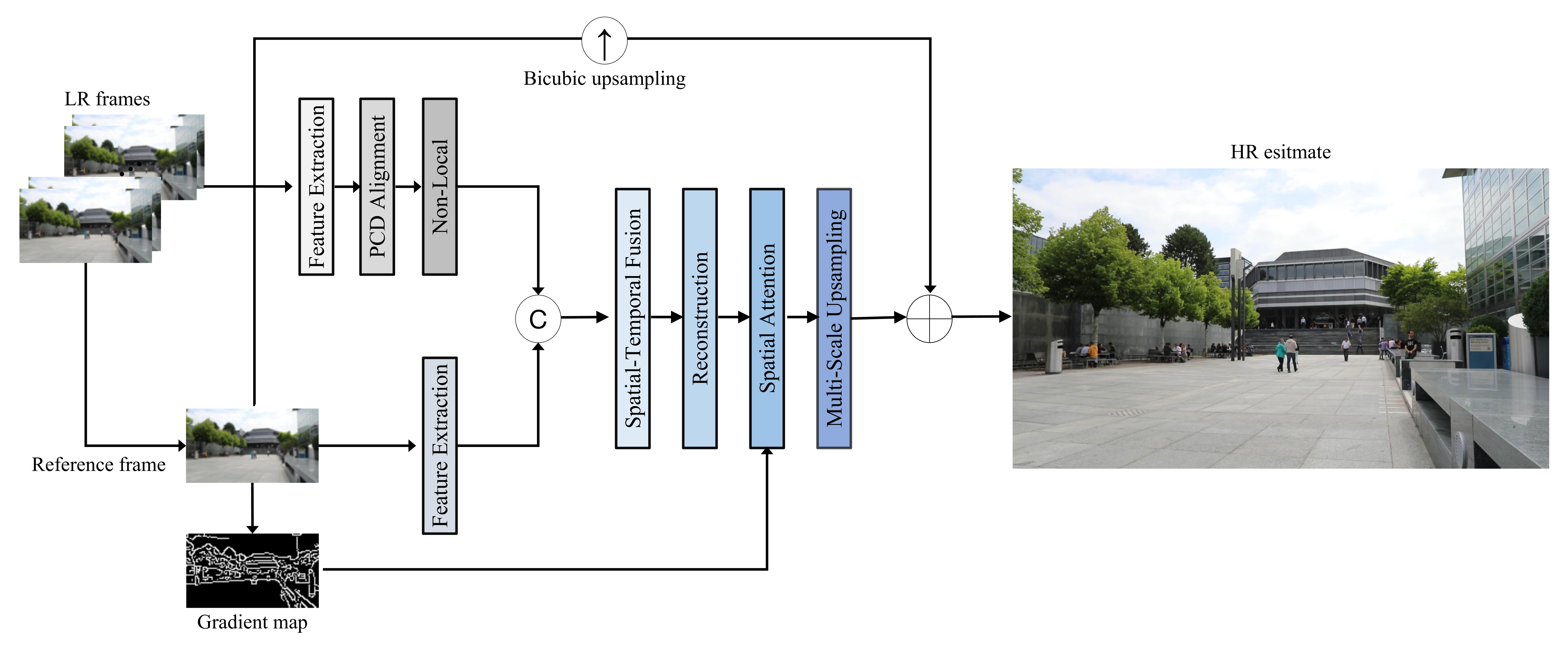}
  \caption{Network proposed by ZZX.}
  \label{fig:zzx_network}
\end{figure}

In order to restore the high-frequency information of the video, the team designed the multi-scale aggregated upsampling based on high frequency attention (MAHA) network. The framework is illustrated in Fig.~\ref{fig:zzx_network}. The team inputs seven LR frames into the feature extraction module, and inspired by the EDVR~\cite{edvr}, the Pyramid, Cascading and Deformable (PCD) alignment module was applied to address global motion. The non-local module was used to select valid inter-frame information. Then, the extracted reference frame features are concatenated with the alignment features that were utilized to perform spatial-temporal fusion in a progressive strategy, which helps to aggregate spatial-temporal information. Next, the team also proposed an attention-guided multi-level residual feature reconstruction module to fully improve feature representation. Finally, to generate a sharp structure HR video, the team computed the gradient map of the LR image to guide the spatial attention module. 
 
The team divided the network training into two stages for both track $1$ and track $2$. For stage $1$, the $L1$ loss was used. For stage $2$, the team fine-tuned the results of stage $1$ and using $0.15*L_{L1} + 0.85*L_{SSIM}$ to train the network. However, the team applied different testing strategies for track $1$ \& $2$ . For Track $1$, they employed model fusion testing and test enhancement strategies to improve the PSNR value. For the track $2$, the best validation performance was used to directly generate the test results.

\subsection{sr\_xxx}

\begin{figure}[t]
  \centering
  \includegraphics[width=\linewidth]{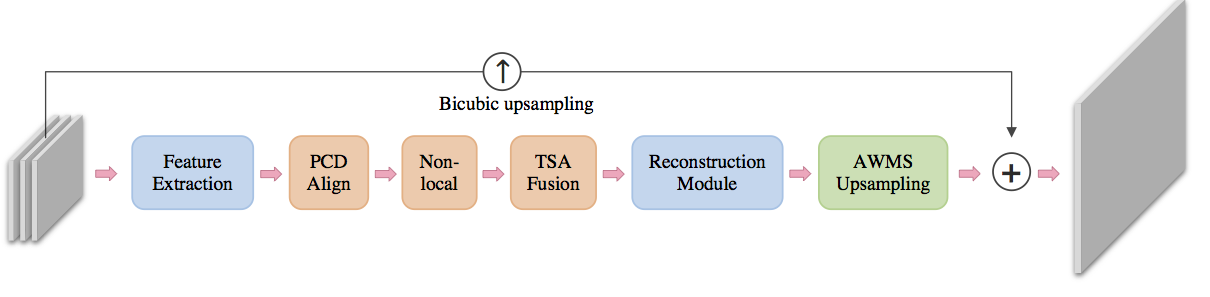}
  \caption{Network proposed by sr\_xxx.}
  \label{fig:srxxx_network}
\end{figure}

The team employs the high-level architecture design of EDVR [1], with improvements to accommodate large upscaling factors with up to 16. The used network is illustrated in Fig.\ref{fig:srxxx_network}. To explain the framework, they first start with EDVR as baseline. EDVR is a unified framework which can achieve good alignment and fusion quality in video restoration tasks. It proposed a Pyramid, Cascading and Deformable (PCD) alignment module,  which for the first time uses deformable convolutions to align temporal frames. Besides, EDVR includes a Temporal and Spatial Attention (TSA) fusion module to emphasize important features.

Their proposed network takes 5 LR frames as input and generates one HR output image frame. They first conduct feature extraction, followed by PCD alignment module, Non-local module and TSA module to align and fuse multiple frames. Right after the reconstruction module, they use adaptive weighted multi-scale (AWMS) module as our upsampling layer. In the last module, they add the learned residual to a direct bicubic upsampled image to obtain the final HR outputs.

They trained two different reconstruction models and ensemble their results to obtain more stable texture reconstructions. To incorporate finer detail ensembling, they combine residual feature aggregation blocks \cite{liu2020residual} and residual channel attention blocks \cite{zhang2018image}.

\subsection{lyl}

\begin{figure}[t]
  \centering
  \includegraphics[width=0.5\linewidth]{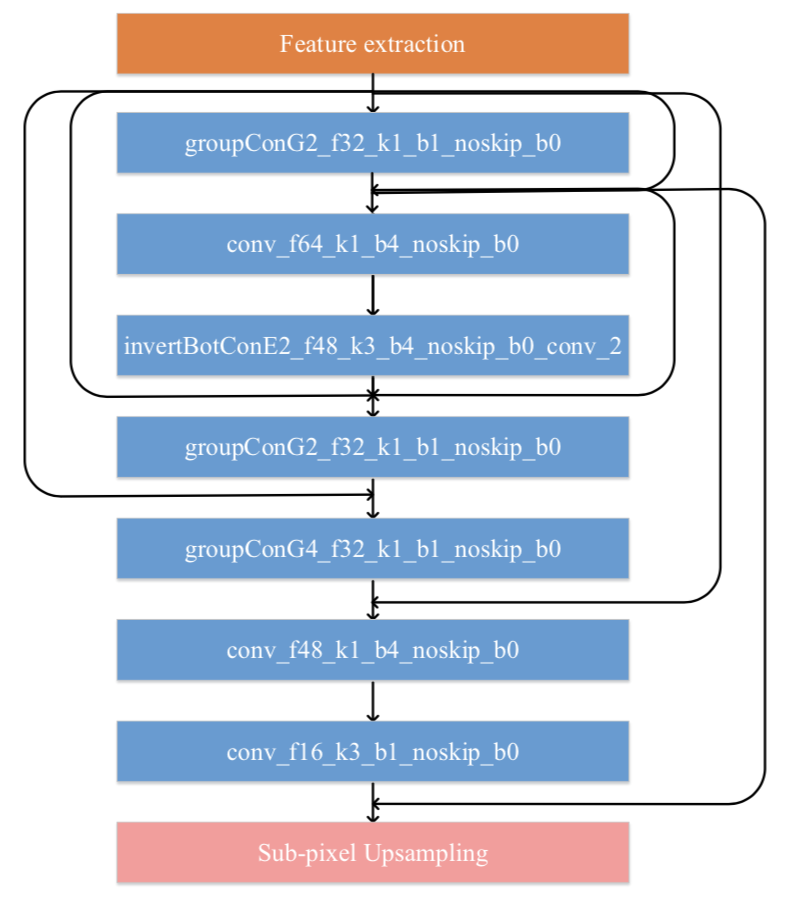} \\
  (a) Coarse Network\\
\includegraphics[width=0.5\linewidth]{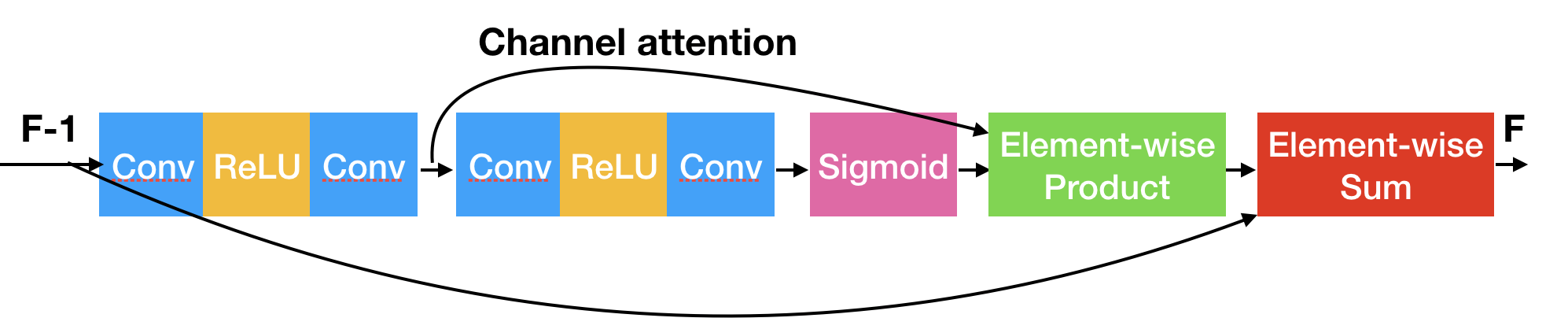} \\
(b) Fine Network

  \caption{ Network proposed by lyl.}
  \label{fig:lyl_network}
\end{figure}

As shown in Fig.\ref{fig:lyl_network}, the team proposes a coarse to fine network for progressive super-resolution reconstruction. By using the suggested FineNet:lightweight upsampling module (LUM), they achieve competitive results with a modest number of parameters. Two requirements are contained in a coarse to fine network(CFN):
(1) progressiveness  and (2) merge the output of the LUM to correct the input in each level. Such a progressive cause-and-effect process helps to achieve the principle for image SR: high-level information can guide an LR image to recover a better SR image. In the proposed network, there are three indispensable parts to enforce the suggested CFN: (1) tying the loss at each level  (2) using LUM structure
and (3) providing a lower level extracted feature input to ensure the availability of low-level information.

They propose to construct their network based on the Laplacian pyramid framework, as shown in Fig.\ref{fig:lyl_network}. Their model takes an LR image as input  and progressively predicts residual images at $S_{1},S_{2}...S_{n}$ levels where $S$ is the scale factor. 
For example, the network consists of 4 sub-networks for super-reconstructing an LR image at a scale factor of 16, if the scale factor is 3, $S=S_{1} \times S_{2}$, $S_{1}=1.5, S_{2}=2$. Their model has three branches: (1) feature extraction and (2) image reconstruction (3) loss function.

\subsection{TTI}

\begin{figure}[t]
  \centering
  \includegraphics[width=0.85\linewidth]{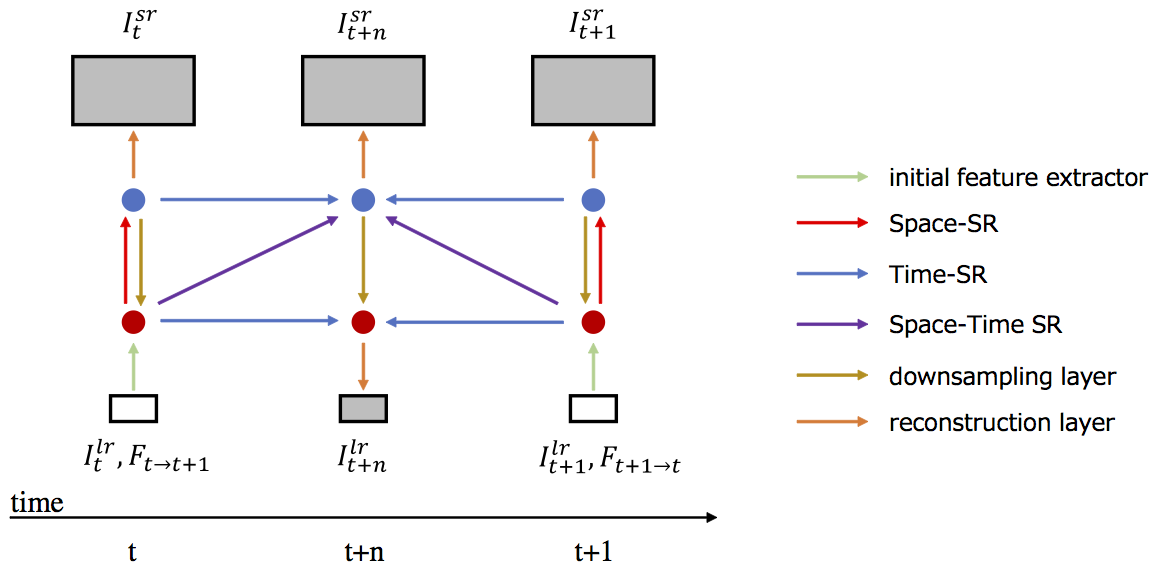}
  \caption{Network proposed by TTI.}
  \label{fig:tti_network}
\end{figure}

The team's base network for x16 video SR is STARnet \cite{haris2020space} shown in Fig.\ref{fig:tti_network}. With the idea that space and time are related, STARnet jointly optimizes three tasks (i.e., space SR, time SR, and space-time SR). In the experiments, STARnet was initially trained using three losses, i.e. space, time, and space-time losses, which evaluate the errors of images reconstructed through space SR paths (red arrows in the figure), time SR paths (blue arrows), and space-time SR paths (purple arrows), respectively. The network is then fine-tuned using only the space loss for optimizing the model, which is specialized for space SR. While space SR in STARnet is basically based on RBPN \cite{haris2019recurrent}, this fine-tuning strategy allows them to be superior to RBPN trained only with the space loss. While the original STARnet employs pyflow \cite{liu2009beyond} for optical flow computation, pyflow almost cannot estimate optical flows in LR frames in this challenge (i.e., 120 $\times$ 67 pixels). The optical flows are too small between subsequent frames. Based on an extensive survey, we chose sift-flow \cite{liu2010sift} that shows better performance on the LR images used in this challenge.

\subsection{CET\_CVLab}

\begin{figure}[t]
  \centering
  \includegraphics[width=0.4\linewidth]{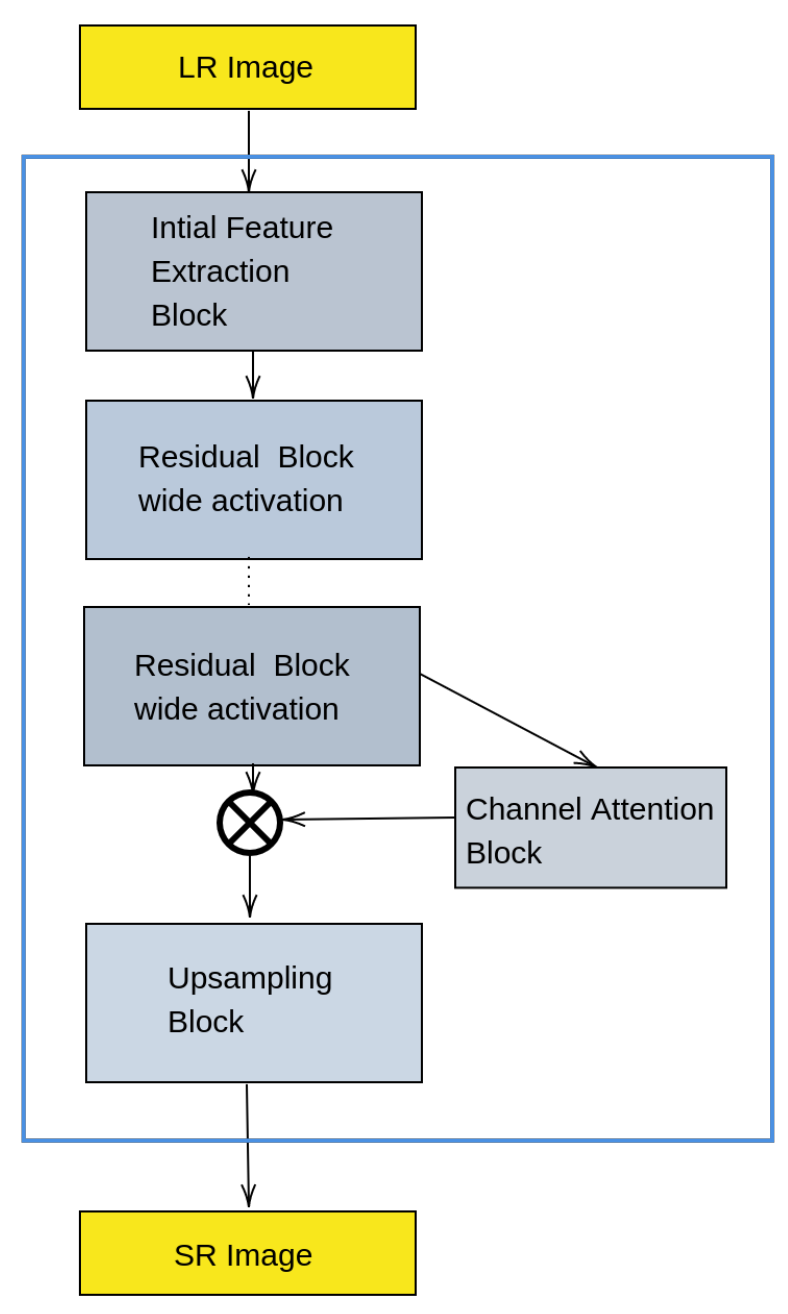}
  \caption{Network proposed by CET-CVLab.}
  \label{fig:cetcvl_network}
\end{figure}

The architecture used by the team is inspired from the wide activation based network in \cite{yu2018wide} and channel attention network in \cite{zhang2018image}. As shown in Fig.\ref{fig:cetcvl_network}, the network mainly consists of 3 blocks. A feature extraction block, a series of wide activation residual blocks and a set of progressive upsampling blocks($\times$2). Charbonnier loss is used for training the network as it better captures the edge information than with mean squared errorloss (MSE).

\section{Challenge Results}

\begin{table*}[t]
    \scriptsize
	\centering
	\newcommand{\sep}{~~}
		\begin{tabular}{lrl|lcccccccc}
			& \phantom{llll} & Method  & $\uparrow$PSNR & $\uparrow$SSIM  & 
    			\begin{tabular}{@{}c@{}} Train  \\ Req \end{tabular}& 
    			\begin{tabular}{@{}c@{}} Train  \\ Time \end{tabular} &
    			\begin{tabular}{@{}c@{}} Test  \\ Req \end{tabular}& 
    			\begin{tabular}{@{}c@{}} Test  \\ Time \end{tabular}& 
    			Params & 
    			\begin{tabular}{@{}c@{}} Extra  \\ Data \end{tabular}\\
			\hline
			\multirow{8}{1mm}{\rotatebox{90}{\resizebox{14mm}{!}{Participants}}}
			& 1. & KirinUK  & \textbf{22.83} & \textbf{0.6450} &   4$\times$V100 & 10d &  1$\times$2080Ti &  6.1s & 45.29M  &  Yes \\ 
			& 2. & Team-WVU  & 22.48 & 0.6378 &  4$\times$TitanXp & 4d &  1$\times$TitanXp &  4.90s & 29.51M  & No \\
			&  & BOE-IOT-AIBD  & 22.48 & 0.6304 &   1$\times$V100 & $>30$d & 1$\times$ 1080 & 4.83s & 53M & No \\
			
			& 4. & sr\_xxx & 22.43 & 0.6353    & 8$\times$V100 & 2d & 1$\times $V100 & 4s & n/a & No \\
			& 5. & ZZX & 22.28 & 0.6321 & 6$\times$1080Ti  & 4d  & 1$\times$1080Ti & 4s & 31,14M  & No \\
			& 6. & lyl & 22.08 & 0.6256 &  1 $\times$ V100 & 2d  & n/a & 13s & n/a  & No \\
			& 7. & TTI & 21.91 & 0.6165 & V100  & n/a  & n/a &  0.249s & n/a & No \\
			& 8. & CET\_CVLab & 21.77 & 0.6112 & 1$\times$P100 &  6d & 1$\times$P100 & 0.04s & n/a & Yes \\
			\hline
		
 			& \multirow{1}{1mm}
			& Bicubic (baseline)  & 20.69 &  0.5770 & & & & & & \\

			\\
		\end{tabular}
	\caption{Quantitative results for track 1. Train Time: days per model, Test Time: seconds per frame.}
	\label{tab:quantitative-results_track1}
\end{table*}


\subsection{Track 1 - Fidelity}

This track aims at restoring the missing high frequencies that were lost during downsampling with the highest fidelity to the underlying ground truth. Challenge track 1 has 65 registered participants, from which 12 submitted solutions to the validation server and 8 teams entered the final ranking. The ranking, along with details about the training and testing are summarized in Tab.~\ref{tab:quantitative-results_track1}. Most provided solutions make use of large networks, as super-resolution with factor 16 is highly challenging and requires high-complexity networks in order to restore the details from learned priors. The top teams mostly employ window based approaches, attention modules and 3D-convolutions to additionally aggregate the temporal information. Team lyl and CET\_CVLab do not process temporal information in their networks and instead rely only on a single frame for the upscaling process. They can not compete with the top teams, which shows the importance of temporal information for high-quality restoration.
The winner in track 1 is team KirinUK with a PSNR score of 22.83dB, followed by team Team-WVU and BOE-IOT-AIBD, which share the second place due to their identical PSNR scores.

\textbf{Metrics}
Since this track is about high fidelity restoration, we rank the teams according to PSNR, which is a pixel-level metric. Additionally, we compute SSIM scores which is a metric based on patch statistics and is considered to correlate better with human perception of image quality. PSNR does not explicitly enforce to retain smooth temporal dynamics. It is therefore possible, that a method can generate high image quality on frame level, but introduces temporal artifacts like flickering. Most window based and 3D-convolution approaches however manage to produce frames with only minimal flickering artifacts, as they have access to adjacent frames. On the other hand, the flickering is very prominent for the single frame enhancers in this challenge.
%

\begin{figure*}[th!]
\begin{center}
{\Large \textbf{Track 1 - Fidelity}\par\medskip}
\includegraphics[width=.95\textwidth]{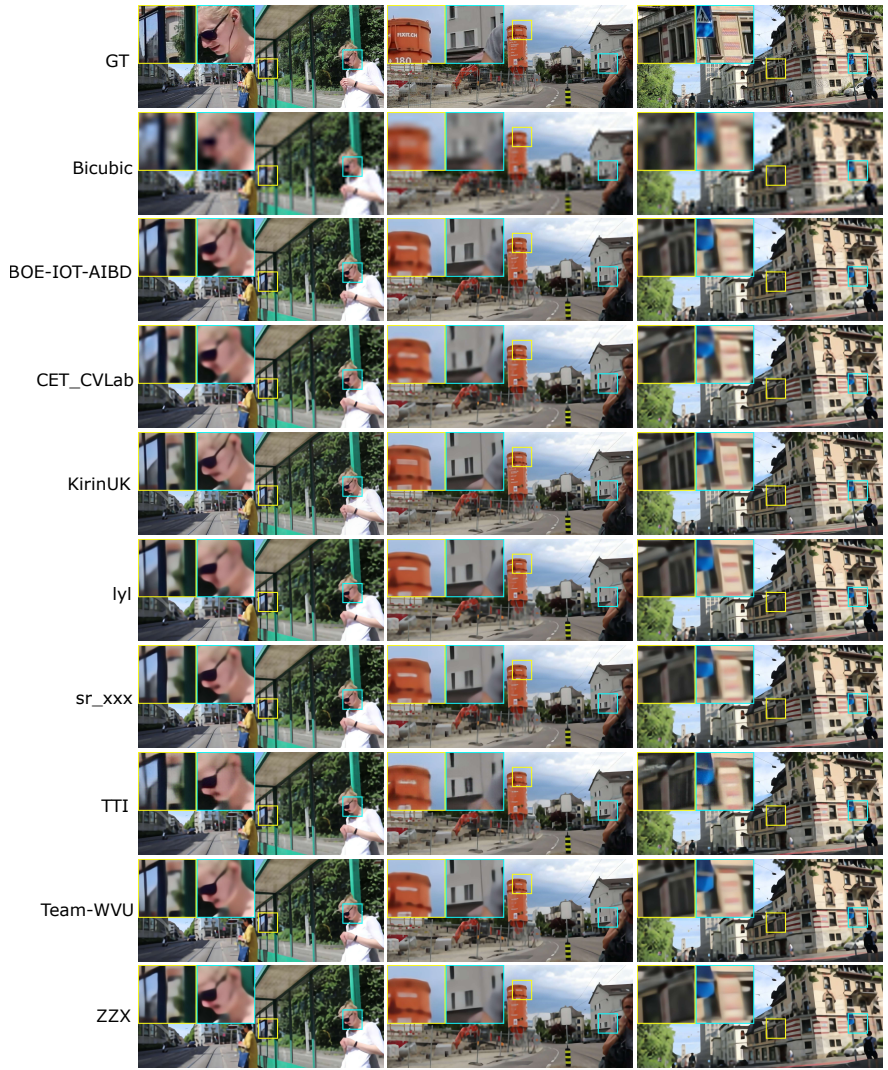}
\end{center}
   \caption{Track 1 Visual results for all competing teams. Additionally, we show the ground truth (GT) and bicubic interpolation (Bicubic) for reference. To present the details more clearly and to fit all methods on a single page, the frames are cropped to $800\times1920$. Highlights from yellow and blue boxes are shown at the top left.}
\label{fig:track1_visual}
\end{figure*}

\textbf{Visual Results}
In addition to the metrics, we also provide visual examples in Fig.~\ref{fig:track1_visual} for all competing methods in the challenge. We also show the Bicubic baseline (MATLAB's \textit{imresize}) together with the ground truth frames for reference. All methods manage to clearly outperform the Bicubic baseline, which is also reflected in the PSNR and SSIM metrics in Tab.~\ref{tab:quantitative-results_track1}. The methods improve PSNR and SSIM by 1.08dB to 2.14dB and 0.0342 to 0.0680 respectively. As expected for such a challenging task, no method is capable of restoring all the fine details present in the ground truth. Predominantly, only sharp edges and smooth textures can be recovered even by the top teams, since the information loss is so extreme. Interestingly, teams KirinUK, Team-WVU, BOE-IOT-AIBD, sr\_xxx, ZZX and TTI restore two windows (blue box highlight, second column) instead of a single one as present in the ground truth frame. This could indicate, the method's upscaling for such a large factor is highly dependent on image priors and having access to temporal information is not sufficient to achieve high restoration quality. Still, a top ranking is only achieved by teams that leverage temporal information, which shows its importance for video super-resolution, even in such extreme settings. 

\subsection{Track 2 - Perceptual}

\begin{table*}[t]
    \scriptsize
	\centering
	\newcommand{\sep}{~~}
		\begin{tabular}{lrl|cccccccc|rr|rrr}
		& \phantom{llll} & Method  & K & T & Z & B (t) & s & B (s) & l & C & 
    			\begin{tabular}{@{}c@{}} Wins  \\ (tot) \end{tabular} &
		    	\begin{tabular}{@{}c@{}} Wins  \\ (\%) \end{tabular} 
			& $\uparrow$PSNR & $\uparrow$SSIM  & $\downarrow$LPIPS  \\
			
			\hline

			\multirow{8}{1mm}{\rotatebox{90}{\resizebox{14mm}{!}{Frame Level}}}
			& 1. & KirinUK  & - & 115 &126& 116 & 113 & 117 & 126 & 144 & \textbf{857} & \textbf{76.52} & \textbf{22.79} & \textbf{0.6474} & \textbf{0.447}\\
			& 2. & Team-WVU &  45 &  - &  85 &  97 &  97 & 110 & 109 & 132 & 675 & 60.27 & 22.48 & 0.6378 & 0.507\\
			& 3. & ZZX &  34 &  75 &   - &  78 &  97 & 101 & 104 & 130 & 619 & 55.27 & 22.09 & 0.6268 & 0.505\\
			& 4. & BOE-IOT-AIBD (t) & 44 &  63 &  82 &   - &  68 &  87 &  95 & 118 & 557 & 49.73 & 21.18 & 0.3633 & 0.514\\
			& 5. & sr\_xxx  & 47 &  63 &  63 &  92 &   - &  95 & 109 & 131 & 600 & 53.57 & 22.43 & 0.6353 & 0.509\\
			&  & BOE-IOT-AIBD (s) & 43 &  50 &  59 &  73 &  65 &   - &  85 & 113 & 488 & 43.57 & 22.48 & 0.6304 & 0.550\\
			& 7. & lyl & 34 & 51 &  56 &  65 &  51 &  75 &    - & 119 & 451 & 40.27 & 22.08 & 0.6256 & 0.535\\
			& 8.& CET\_CVLab & 16 & 28 &  30 &  42 &  29 &  47 &  41 &   - & 233 & 20.80 & 21.77 & 0.6112 & 0.602 \\[0.1cm]

			&&&&&&&&&&&&&\multicolumn{3}{c}{Final Scores}\\
			&&&&&&&&&&&&& Frame & Video & Total \\
			\hline
			\multirow{8}{1mm}{\rotatebox{90}{\resizebox{14mm}{!}{Video Level}}}
			& 1. & KirinUK  & - & 7 & 7 & 8 & 7 & 8 & 8 & 6 & \textbf{51} & \textbf{72.86} & 76.52 & 72.86 &\textbf{149.38}\\
			& 2. & Team-WVU &  3 & - & 7 & 6 & 8 & 2 & 8 & 8 & 42 & 60.00 &  60.27 & 60.00 & 120.27\\
			& 3. & ZZX & 3 & 3 & - & 5 & 5 & 4 & 9 & 8 & 37 & 52.86 & 55.27 & 52.86 & 108.13\\
			& 4. & BOE-IOT-AIBD (t) & 2 & 4 & 5 & - & 8 & 8 & 5 & 8 & 40 & 57.14 & 49.73 & 57.14 & 106.87  \\
			& 5. & sr\_xxx & 3 & 2 & 5 & 2 & - & 4 & 7 & 6 & 29 & 41.43 & 53.57 & 41.43 & 95.00\\
			&  & BOE-IOT-AIBD (s) & 2 & 8 & 6 & 2 & 6 & - & 5 & 7 & 36 & 51.43 & 43.57 & 51.43 & 95.00\\
			& 7. & lyl & 2 & 2 & 1 & 5 & 3 & 5 & - & 7 & 25 & 35.71 & 40.27 & 35.71 & 75.98\\
			& 8. & CET\_CVLab & 4 & 2 & 2 & 2 & 4 & 3 & 3 & - & 20 & 28.57 & 20.80 & 28.57 & 49.37\\\\
		\end{tabular}
	\caption{User study results for track 2. The results are obtained by a one vs. one user study on frame level and video level. Wins (tot) indicates absolute wins in all comparisons. Wins (\%) reflects relative wins, which are normalized by the number of comparisons with other teams. Compared to absolute wins, the relative wins allow direct comparison between frame level and video level performance. The aggregated relative wins of both studies on frame and video level led to the final ranking. Additionally, we provide PSNR, SSIM and LPIPS scores for reference. Note, these metrics are not considered for ranking.}
	\label{tab:quantitative-results_track2}
\end{table*}

\begin{figure}[h]
  \centering
  \includegraphics[width=\linewidth]{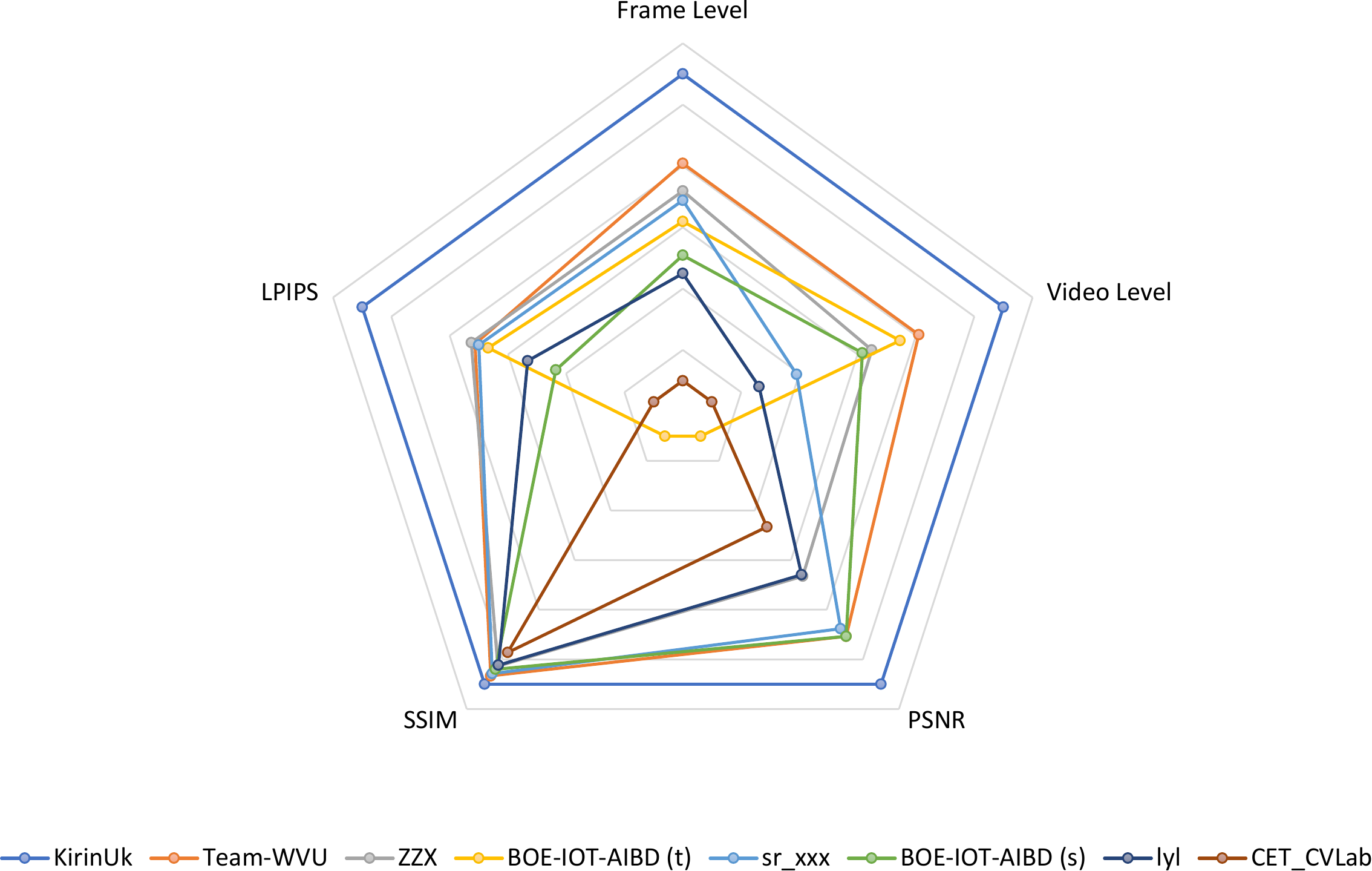}
  \caption{Radar plot for track 2. All values per axis have been normalized to lie in [0,1]. The range for LPIPS has been reversed to indicate better values towards the outside in accordance with the other metrics.}
  \label{fig:radar_plot_track2}
\end{figure}

Due to the extreme information loss, it is hard to accurately restore the high-frequency content with respect to the ground truth. If deviations from the ground truth can be accepted and more visually pleasing results are desired, perceptual quality can be traded-off for fidelity. The results may not entirely reflect the underlying ground truth, but instead boost the perceptual quality considerably. We therefore do not rely on PSNR and SSIM for evaluation in track 2, but instead conduct a user study to asses human perceptual quality. Challenge track 2 has 54 registered participants, from which 7 submitted solutions to the validation server and 7 teams entered the final ranking

\textbf{Metrics}
Assessing perceptual quality quantitatively is difficult and remains largely an open problem. Attempts for such metrics have been made in the past and one of the most promising metrics is called Learned Perceptual Image Patch Similarity (LPIPS), which is proposed in \cite{lpips}. This metric measures similarity to the ground truth in feature space of popular architectures, e.g. \cite{alexnet}. 
While this metric is widely adopted for perceptual quality assessment, especially on images, it still fails in some cases to reflect human perception. Like PSNR and SSIM, it does not discriminate on  temporal dynamics, which are crucial for high quality videos.

\textbf{User Study}
Since quantitative metrics are not reliable, we resort to a user study to rank the participating teams in track 2. For that matter we split the evaluation in two separate user studies, a frame level study and a video level study. The frame level study is meant to judge the image level quality and is performed on randomly subsampled frames from all 16 sequences in the test set. The competing method's frames are compared side-by-side in a one vs. one setting, resulting in 28 comparisons per frame. We asked 10 users to judge the frame level quality, which results in $16\times28\times10=4480$ total ratings. The detailed results are shown in a confusion matrix in Tab.~\ref{tab:quantitative-results_track2}. Each row shows the preference of the method in the first column against all other methods. Additionally, we show the total number of preferences (Wins (tot)) plus the relative preferences, which are normalized to the number of total comparisons $16\times7\times10=1120$ with all teams (Wins (\%)). 
The video level study is meant for evaluating the temporal dynamics in the videos and the overall perceptual quality when watching the videos. Again, we generated side-by-side videos between all methods for comparison in a one vs. one setting. The videos are generated by compiling all 1920 frames from 16 sequences into a single video, showing two competing methods. This results in 28 short videos of $\approx$~1~minute. We also ask 10 users to perform the video level user study and get a total number of $28\times10=280$ ratings. In order to directly compare with the frame level study, we normalize the total wins by the total comparisons with all the teams ($10\times7=70$) to get the relative scores. We derive the ranking from the combined relative scores of both frame level and video level user studies (see Tab.~\ref{tab:quantitative-results_track2}, lower right). Team KirinUk is the clear winner in track 2, followed by Team-WVU and ZZX on the second and third place.
We also provide PSNR, SSIM and LPIPS metrics for reference. A qualitative illustration of all track 2 results is presented in Fig.~\ref{fig:radar_plot_track2}, including the user study results. Note, the final ranking is only derived from the user study.

\begin{figure*}[th!]
\begin{center}
{\Large \textbf{Track 2 - Perceptual}\par\medskip}
\includegraphics[width=.95\textwidth]{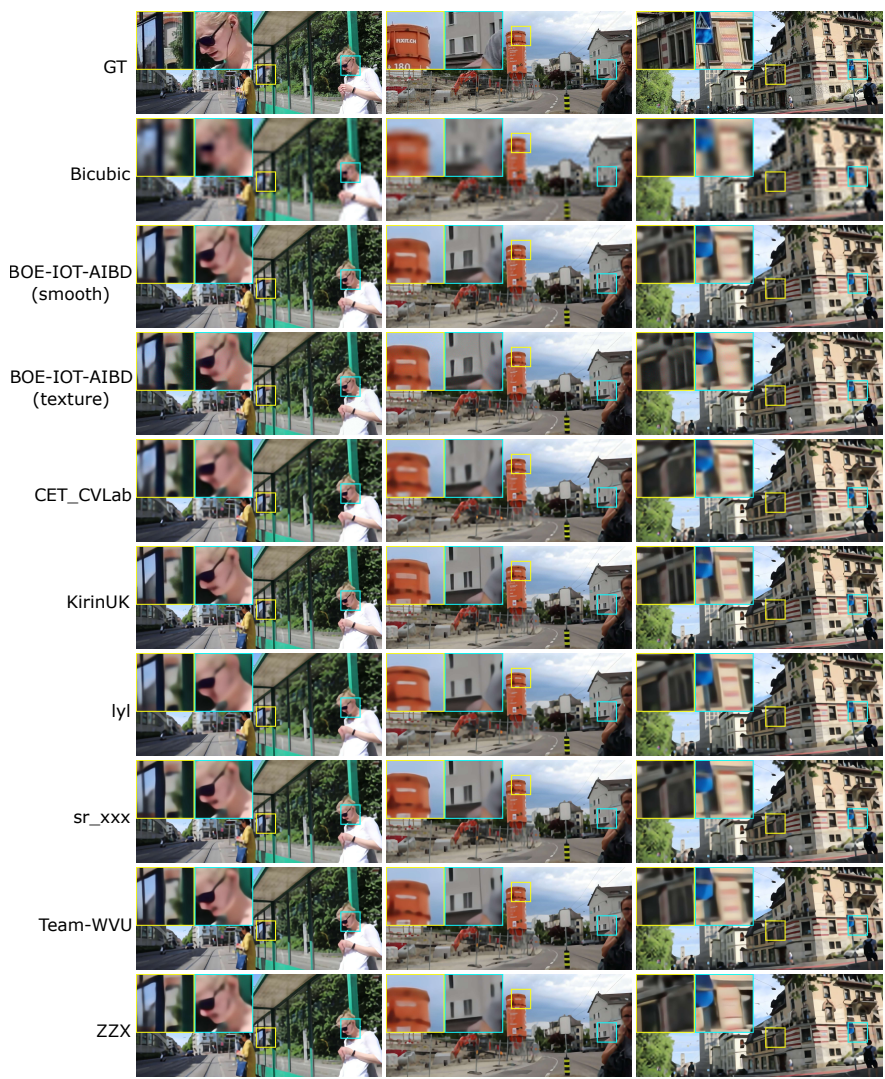}
\end{center}
   \caption{Track 2 Visual results for all competing teams. Additionally, we show the ground truth (GT) and bicubic interpolation (Bicubic) for reference. To present the details more clearly and to fit all methods on a single page, the frames are cropped to $800\times1920$. Highlights from yellow and blue boxes are shown at the top left. Team BOE-IOT-AIBD provides two distinct solutions, which focus on high quality textures and temporal smoothness respectively.}
\label{fig:track2_visual}
\end{figure*}

\textbf{Visual Results}
To allow direct comparison to track 1, we provide the visual results on the same frames for track 2 in Fig.~\ref{fig:track2_visual}. Note that teams Team-WVU, sr\_xxx, lyl and CET\_CVLab submitted the same set of frames to both tracks, while KirinUK, BOE-IOT-AIBD and ZZX adapted their solutions to the specific requirements in both tracks. BOE-IOT-AIBD even provided two distinct solutions for track 2. One is optimized with emphasis on textures, the other is designed for temporal smoothness, abbreviated with (t) and (s) respectively in Tab.~\ref{tab:quantitative-results_track2}. Surprisingly, the texture based solution of BOE-IOT-AIBD also performs better in the video level user study. According to the users, the sharper texture details seem to have a higher impact on the quality than the flickering artifacts. The winning team KirinUK manages to not only outperform all other teams in both user studies, but also in the provided metrics PSNR, SSIM and LPIPS. However, it has to be considered, that the solution is optimized with L1 and VGG-loss, which are both closely related to these metrics. BOE-IOT-AIBD and KirinUK are the only teams that incorporate a GAN loss into their training strategy. On the other hand, Team-WVU trains its network only on the pixel-based Charbonnier Loss, and ZZX trains their perceptual solution on L1 and SSIM. Nevertheless, they outperform BOE-IOT-AIBD, which employs a GAN loss. Therefore, strong guidance from a pixel-based loss might be important for such an extreme scaling factor.

\section{Conclusions}
This paper presents the AIM 2020 challenge on Video Extreme Super-Resolution. We evaluate the performance in this challenging setting for both high fidelity restoration (track 1) and perceptual quality (track 2). The overall winner KirinUK manages to strike the best balance between restoration and perceptual quality. The participating teams provided innovative and diverse solutions to deal with the extreme upscaling factor of 16. Further improvements could be achieved by reducing and ideally removing the notorious flickering artifacts associated with video enhancement in general. On top of that, a more powerful generative setting could be designed for higher perceptual quality in track 2. Quantitative evaluation for perceptual quality still requires more research, especially in the video domain, where temporal consistency is important. We hope this challenge attracts more researchers to enter the area of extreme video super-resolution as it offers great opportunities for innovation.

\section*{Acknowledgements}
We thank the AIM 2020 sponsors: Huawei, MediaTek, NVIDIA, Qualcomm, Google, and Computer Vision Lab (CVL), ETH Zurich.

\section*{Appendix A: Teams and Affiliations}
\label{sec:affiliation}


\noindent
{\large{\textbf{AIM 2020 team}}}

\noindent
\textit{\textbf{Title:}} AIM 2020 Challenge on Video Extreme Super-Resolution

\noindent
\textit{\textbf{Members: }}Dario Fuoli, Zhiwu Huang, Shuhang Gu, Radu Timofte

\noindent
\textit{\textbf{Affiliations:}}

\noindent
Computer Vision Lab, ETH Zurich, Switzerland; The University of Sydney, Australia \\

\noindent
{\large{\textbf{KirinUK}}}

\noindent
\textit{\textbf{Title:}} Efficient Video Enhancement and Super-Resolution Net (EVESRNet)

\noindent
\textit{\textbf{Members: }}
Arnau Raventos,
Aryan Esfandiari,
Salah Karout

\noindent
\textit{\textbf{Affiliations:}}

\noindent
Huawei Technologies R\&D UK
\\

\noindent
{\large{\textbf{Team-WVU}}}

\noindent
\textit{\textbf{Title:}} Multi-Frame based Deformable Kernel Convolution Networks

\noindent
\textit{\textbf{Members: }}
Xuan Xu,
Xin Li,
Xin Xiong,
Jinge Wang

\noindent
\textit{\textbf{Affiliations:}}

\noindent
West Virginia University, USA;
Huazhong University of Science and Technology, China
\\

\noindent
{\large{\textbf{BOE-IOT-AIBD}}}

\noindent
\textit{\textbf{Title:}} Fully 3D–Convolutional MultiGrid–BackProjection Network

\noindent
\textit{\textbf{Members: }}
Pablo Navarrete Michelini,
Wenhao Zhang

\noindent
\textit{\textbf{Affiliations:}}

\noindent
BOE Technology Group Co., Ltd
\\

\noindent
{\large{\textbf{ZZX}}}

\noindent
\textit{\textbf{Title:}} Multi-Scale Aggregated Upsampling Extreme Video Based on High Frequency Attention

\noindent
\textit{\textbf{Members: }}
Dongyang Zhang, Hanwei Zhu, Dan Xia

\noindent
\textit{\textbf{Affiliations:}}

\noindent
Jiangxi University of Finance and Economics; National Key Laboratory for Remanufacturing, Army Academy of Armored Forces
\\

\noindent
{\large{\textbf{sr\_xxx}}}

\noindent
\textit{\textbf{Title:}} Residual Receptive Attention for Video Super-Resolution

\noindent
\textit{\textbf{Members: }}
Haoyu Chen,
Jinjin Gu,
Zhi Zhang

\noindent
\textit{\textbf{Affiliations:}}

\noindent
Amazon Web Services; The Chinese University of Hong Kong, Shenzhen
\\

\noindent
{\large{\textbf{lyl}}}

\noindent
\textit{\textbf{Title:}} Coarse to Fine Pyramid Networks for Progressive Image Super-Resolution

\noindent
\textit{\textbf{Members: }}
Tongtong Zhao,
Shanshan Zhao

\noindent
\textit{\textbf{Affiliations:}}

\noindent
Dalian Maritime University;
China Everbright Bank Co., Ltd
\\

\noindent
{\large{\textbf{TTI}}}

\noindent
\textit{\textbf{Title:}} STARnet

\noindent
\textit{\textbf{Members: }}
Kazutoshi Akita, Norimichi Ukita

\noindent
\textit{\textbf{Affiliations:}}

\noindent
Toyota Technological Institute (TTI)
\\

\noindent
{\large{\textbf{CET\_CVLab}}}

\noindent
\textit{\textbf{Title:}} Video Extreme Super-Resolution using Progressive Wide Activation Net

\noindent
\textit{\textbf{Members: }}
Hrishikesh P S,
Densen Puthussery,
Jiji C V

\noindent
\textit{\textbf{Affiliations:}}

\noindent
College of Engineering, Trivandrum, India
\\

%
%
\bibliographystyle{splncs04}
\bibliography{egbib}
\end{document}